\documentclass[letterpaper, 10 pt, conference]{ieeeconf}  

\IEEEoverridecommandlockouts                              

\usepackage{graphicx} 
\usepackage[colorlinks=true]{hyperref}

\title{\LARGE \bf
Domain-Specific Pre-training Improves Confidence in Whole Slide Image Classification
}

\author{\authorblockN{Soham Rohit Chitnis$^{1}$, Sidong Liu$^{2,*}$, Tirtharaj Dash$^{3}$, Tanmay Tulsidas Verlekar$^{1}$, \\Antonio Di Ieva$^{2,*}$, Shlomo Berkovsky$^{2,*}$, Lovekesh Vig$^{4}$, Ashwin Srinivasan$^{1}$}
\thanks{*Sidong Liu and Antonio Di Ieva were supported by an Australian NHMRC Ideas Grant.}%
\thanks{$^{1}$Soham Rohit Chitnis, Tanmay Tulsidas Verlekar, and Ashwin Srinivasan are with Department of Computer Science \& Information Systems, Birla Institute of Technology and Science, Pilani}
\thanks{$^{2}$Sidong Liu, Antonio Di Ieva and Shlomo Berkovsky are with Faculty of Medicine, Health and Human Sciences, Macquarie University}
\thanks{$^{3}$Tirtharaj Dash is with Boolean Lab, University of California, San Diego}
\thanks{$^{4}$Lovekesh Vig is with TCS Research, India}
}
\DeclareUnicodeCharacter{2009}{ }

\begin{document}

\maketitle
\thispagestyle{empty}
\pagestyle{empty}
\bstctlcite{IEEEexample:BSTcontrol}

\begin{abstract}

Whole Slide Images (WSIs) or histopathology images are used in digital pathology. WSIs pose great challenges to deep learning models for clinical diagnosis, owing to their size and lack of pixel-level annotations. With the recent advancements in computational pathology, newer multiple-instance learning-based models have been proposed. Multiple-instance learning for WSIs necessitates creating patches and uses the encoding of these patches for diagnosis. These models use generic pre-trained models (ResNet-50 pre-trained on ImageNet) for patch encoding. The recently proposed KimiaNet, a DenseNet121 model pre-trained on TCGA slides, is a domain-specific pre-trained model. This paper shows the effect of domain-specific pre-training on WSI classification. To investigate the effect of domain-specific pre-training, we considered the current state-of-the-art multiple-instance learning models, 1) CLAM, an attention-based model, and 2) TransMIL, a self-attention-based model, and evaluated the models' confidence and predictive performance in detecting primary brain tumors - gliomas. Domain-specific pre-training improves the confidence of the models and also achieves a new state-of-the-art performance of WSI-based glioma subtype classification, showing a high clinical applicability in assisting glioma diagnosis. We will publicly share our code and experimental results at 
\url{https://github.com/soham-chitnis10/WSI-domain-specific}.
\newline

\end{abstract}

\section{Introduction}
\label{sec:intro}

Currently, histopathology is the clinical gold standard for tumor assessment and diagnosis. With advancements in digital pathology and artificial intelligence (AI), deep learning techniques have shown great potential in assisting tumor diagnosis. However, deep learning-based approaches require either manually annotated whole slide images (WSIs) or large datasets with slide-level labels in a weakly supervised setting. The slide-level labels may correspond to small regions from the gigapixel WSIs. Therefore, most of the approaches rely on pixel, patch, or regions-of-interest (ROI) level annotations \cite{Wang2019,Chen2019,Nagpal2019,Bejnordi2017}. There have been attempts to assign the same label to each patch \cite{Coudray2018}, but these approaches suffer from noisy training and are not applicable when the slide-level label corresponds to tiny regions from the image. 

Recently proposed multiple-instance learning models use feature embedding \cite{pmlr-v80-ilse18a,10.1001/jamanetworkopen.2019.14645,DBLP:journals/corr/abs-2001-01599,Naik2020,Lu2021,Shao2021TransMILTB}. Clustering-constrained Attention Multiple instance learning (CLAM) \cite{Lu2021} is a deep learning-based digital pathology framework for weakly supervised WSI classification using attention-based instance-level clustering. The CLAM algorithm does not require ROI extraction, or pixel or tile or patch-level annotations. The algorithm works based on the attention scores assigned to the tiles or patches during the training, where the highest attention scores are considered positive evidence of the class. A ResNet50 network \cite{DBLP:journals/corr/HeZRS15}  pre-trained on the ImageNet dataset is used for extracting features from tiles or patches obtained from the segmented tissue region. CLAM produces interpretable heatmaps that allow medical practitioners to visualize the tissue regions that contribute to the predictions of the model. These heatmaps displaying the morphological features can be verified by trained pathologists whether the decisions made by the model align with the manual diagnostic determinations. TransMIL \cite{Shao2021TransMILTB} is a recently proposed model which is based on a transformer model. Just like CLAM, TransMIL only requires a slide-level label and also shows great interpretability. 

Training a deep neural network from scratch for specific problem statements may not be feasible due to limited available data, or training may take a very long time. Therefore, many researchers exploit pre-trained models. In domain-specific pre-training, the models are first pre-trained on a large dataset, followed by a fine-tuning over the domain-specific dataset. Generally, in computer vision, models pre-trained on ImageNet data have been used for domain-specific downstream tasks. We call a model pre-trained on ImageNet as generic pre-trained, and the model fine-tuned on a domain-specific dataset as domain-specific pre-trained.

Current state-of-the-art WSI classification models, such as CLAM and TransMIL, are based on features extracted by generic pre-trained models, e.g., ResNet50 pre-trained on ImageNet samples, which differ from WSIs in terms of color, texture, morphological, and geometric representations. However, the impact of domain-specific pre-training on WSI classification has not yet been investigated. Our main contributions are as follows:
\begin{itemize}
    \item We introduce Confidence as an evaluation metric for WSI classification.
    \item We investigate the impact of domain-specific pre-training on the model's performance and confidence.
\end{itemize}


\section{Empirical Evaluation}
\label{sec:empeval}

\subsection{Aims}
\label{sec:aims}

We conjecture that domain-specific pre-training improves not just the accuracy, but also the \textit{confidence} in prediction of the models for WSI classification. We will test this conjecture using the following pre-training methods (a) a generic pre-trained model, DenseNet121 (pre-trained on ImageNet Data), (b) a domain-specific pre-trained model, KimiaNet (pre-trained on WSIs). Generic and domain-specific pre-trained models are used as inputs to two different approaches for constructing models for WSI: (a) CLAM and (b) TransMIL.

\subsection{Datasets}
\label{sec:mat}

Two datasets are used for pre-training feature extractors, including ImageNet and The Cancer Genome Atlas (TCGA). ImageNet dataset is a classical collection of natural images for the visual recognition task \cite{Russakovsky2015}. ResNet50 and DenseNet121 Networks are pre-trained on this dataset. TCGA is the large domain-specific dataset on which KimiaNet was trained. This is a publicly available repository \cite{Gutman2013,Tomczak2015,Cooper2018} with 30,072 WSIs. KimiaNet was trained on a subset of this dataset with 11,579 WSIs with permanent hematoxylin and eosin (H\&E) sections. Further details are available in \cite{Riasatian2021}.

The Digital Brain Tumor Atlas (DBTA) dataset \cite{Roetzer-Pejrimovsky2022} is used to train and test the classification models. DBTA dataset consists of 3,115 slides of 126 brain tumor types. A subset of this dataset contains a total of 866 histopathology slides of 791 patients with the 5 primary brain tumor types with molecular subtypes based on the 2021 WHO classification of central nervous system tumors \cite{Louis2021}, including 117 astrocytoma-IDH mutant, 66 astrocytoma-IDH wildtype, 176 oligodendroglioma-IDH wildtype, 34 glioblastoma-IDH mutant, and 473 glioblastoma-IDH wildtype WSIs. The 70:15:15 split is used for the training, validation, and test sets. As several patients have multiple slides, we ensure that different slides of a patient do not exist in both the training and test sets.

\subsection{Models}

We focus on the recently developed state-of-the-art models, CLAM \cite{Lu2021}, and TransMIL \cite{Shao2021TransMILTB}, both of which are multiple-instance learning models. \\

\textbf{CLAM}: We consider two variants of CLAM, namely single branch (CLAM-SB) and multiple branches (CLAM-MB). CLAM uses attention-based pooling \cite{pmlr-v80-ilse18a} to aggregate the slide-level representations from patch-level representations. CLAM has N multi-class attention branches for the N multi-classification problem, which are used to score the class-specific slide-level representations. During training, it has the instance clustering level, which can learn the class-specific features; therefore, CLAM uses a weighted loss of the slide-level classification, and the instance-level clustering loss is used. The slide-level classification is the standard cross-entropy loss, and the instance clustering loss is a smooth SVM loss. For the instance clustering level, we consider $B=8$, which is 8 positive and negative samples. \\

\textbf{TransMIL} It uses a self-attention-based mechanism to model the correlation between the patches. TransMIL has two transformer layers for aggregating the morphological information. TransMIL has adopted the approximation for self-attention proposed in the Nystr{\"o}m method \cite{DBLP:journals/corr/abs-2102-03902}. It uses the Pyramid Positional Encoding Generator (PPEG) for positional encoding, which can encode global and context information. TransMIL uses the standard cross-entropy loss for training the model. 

\subsection{Evaluation Metrics}
The state-of-the-art models have used accuracy and Area Under the Receiver Operating Characteristic Curve (AUC) as the evaluation metrics for benchmarking. We also evaluate the confidence of models, which is the average class probability of the predicted class. Confidence is given by the equation (\ref{eq:conf}), where $y_{pred}$ is the predicted class-label, and $x$ is a sample from the test set $X$.
\begin{equation}
\small Confidence = \frac{\sum_{x} P(y_{pred}| x)}{|X|}
\label{eq:conf}
\end{equation}
\subsection{Algorithms and Machines}
We train all models on NVIDIA Tesla V100-SXM2 GPU with 32GB memory. The code has been implemented using PyTorch Deep Learning Framework, and Python \textit{net:cal} library \cite{Kueppers_2020_CVPR_Workshops} is used for evaluation of the confidence of the model.

\section{Method}
\label{sec:method}

\subsection{Pre-Processing}

\begin{figure}
    \centering
    \includegraphics[width=0.95\columnwidth,height=0.38\textheight]{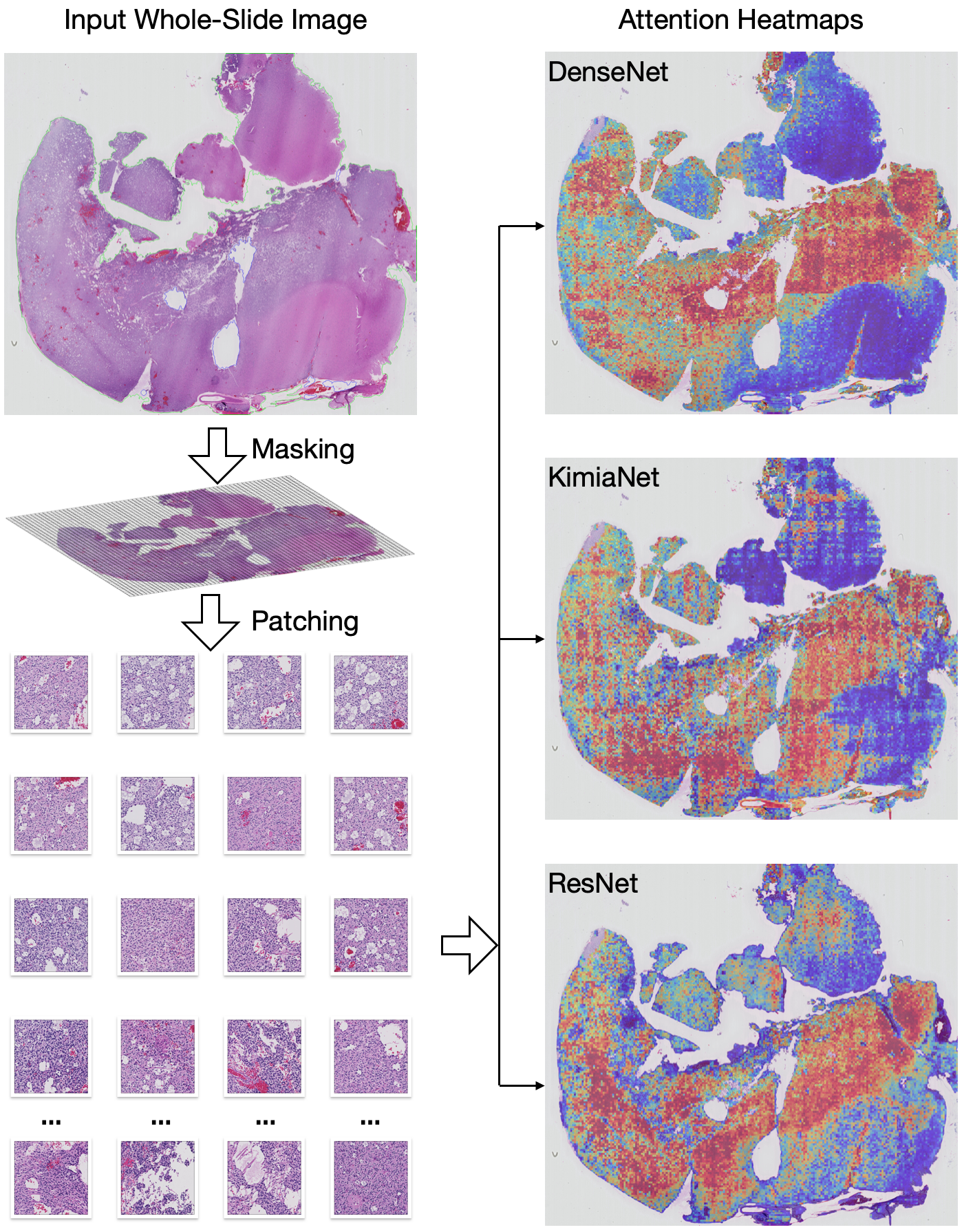}
    \caption{The WSI pre-processing pipeline.}
    \label{fig:preprocessing}
\end{figure}

The gigantic size of WSIs poses challenges in training a model in an end-to-end fashion and would require expensive computing power; therefore, multiple-instance learning models are proposed. In the context of WSI classification, we consider the slide as the bag and the patches of the slides as the instances. Figure \ref{fig:preprocessing} shows the key steps of the WSI pre-processing pipeline. \\

\textbf{Segmentation and patching}
 For the segmentation and patching of each slide, we use the highly efficient method proposed by CLAM \cite{Lu2021}. Patches of size 256 x 256 are cropped from the segmented foreground contours. DBTA has the slide image captured at 20x and 40x resolution. The higher the magnification, the higher the number of patches, leading to higher computational requirements. Therefore all models have been trained at 20x magnification.\\

\textbf{Feature extraction}
The MIL-based models require creating patches from the slides. CLAM \cite{Lu2021} and TransMIL \cite{Shao2021TransMILTB} methods require a feature extraction process. They extract the features using a generic pre-trained model, ResNet50 network pre-trained of the ImageNet data. Therefore, ResNet50 is considered the baseline. DenseNet architecture has shown better results on the ImageNet dataset due to the dense connections between all layers with matching feature-maps sizes \cite{DBLP:journals/corr/HuangLW16a}. Instead of the generic pre-trained model (DenseNet121), we propose to use the domain-specific pre-trained network called KimiaNet \cite{Riasatian2021}, which is a DenseNet121 network pre-trained on TCGA data. We will consider the DenseNet121 network, pre-trained on ImageNet, for comparison with KimiaNet, whose structure is the same.

\subsection{Experimental Procedure}

Our procedure to test the experimental conjecture in Sec. \ref{sec:aims} is straightforward:

\begin{enumerate}
\item Let a slide be denoted by $\mathit{s}$, where $\mathit{s}$ $\in$ $\mathit{S}$, the set of slides
\item For each $\mathit{s}$ $\in$ $\mathit{S}$, segment and create patches; the set of all patches of $\mathit{s}$ is denoted by $\mathit{p}$
\item For each pre-training method in \{ResNet50, DenseNet121, KimiaNet\}
    \begin{itemize}
        \item For each $\mathit{p}$, extract features
        \item Let $\mathit{D}$ denote an array of data seeds, and $\mathit{M}$ denote an array of model seeds
        \item For each $data$-$seed$ in $\mathit{D}$:
        \begin{itemize}
            \item Split the dataset into train/validation/test sets
            \item For each $model$-$seed$ in $\mathit{M}$:
            \begin{itemize}
                \item Use the selected features to train each classification model in $\{$CLAM-SB, CLAM-MB, TransMIL$\}$
                \item Estimate the test accuracy, AUC, and confidence of the classification model
            \end{itemize}
        \end{itemize}
    \end{itemize}
\end{enumerate}

The following additional details are relevant:
\begin{itemize}
\item[-] $data$-$seed$ is responsible for the train/validation/test split, and $model$-$seed$ is responsible for the initialization of model parameters. The number of $data$-$seed$ is 5 and the number of $model$-$seed$ is 3 in this study, so there are 15 combinations in total. In all cases, average values over all 15 experiments of accuracy, AUC, and confidence are used for comparison. 

\item[-] We train models for 200 epochs in each run and use `early stopping' with patience of 20 epochs. We train the model for at least 50 epochs to ensure the model has converged. The learning rate is set to 2e-4 with weight decay of 1e-5. The batch size is set to 1. 

\item[-] Adam optimizer \cite{Kingma2014AdamAM} is used for training CLAM, and Lookahead optimizer \cite{10.5555/3454287.3455148} is used for training TransMIL.

\item[-] Multinomial sampling is used to mitigate the class imbalance problem in the training set. The multinomial sampling probabilities are inversely proportional to the frequency of the ground truth.

\end{itemize}

\section{Results}
\label{sec:results}

The main results from our experiments comparing generic and domain-specific pre-training are presented in Table \ref{tab:main-results}. The most interesting observation is that domain-specific pre-training results in higher average confidence in prediction (by $0.3\%$ to $1.3\%$), which appears to be less well-known. The impact of domain-specific pre-training on the model's test accuracy and AUC are not conclusive. For example, when using CLAM as the classification model, KimiaNet results in higher accuracy, but DenseNet results in higher AUC, and it is the opposite when using TransMIL. We also note that, unlike the results in \cite{Shao2021TransMILTB}, which showed a clear advantage of using the transformer-based approach, we find results for TransMIL are mixed. This may be due to differences in the dataset we have used, which is different to the benchmarks used, or due to differences in hyper-parameter tuning. 


\begin{table}[!htb]
\centering
\caption{Evaluation of Domain-specific Pre-training}
\label{tab:main-results}
\resizebox{\columnwidth}{!}{%
\begin{tabular}{|c|cc|cc|cc|}
\hline
 & \multicolumn{2}{c|}{AUC} & \multicolumn{2}{c|}{Accuracy} & \multicolumn{2}{c|}{Confidence} \\ \hline
Models & \multicolumn{1}{c|}{DenseNet} & KimiaNet & \multicolumn{1}{c|}{DenseNet} & KimiaNet & \multicolumn{1}{c|}{DenseNet} & KimiaNet \\ \hline
CLAM-SB & \multicolumn{1}{c|}{\textbf{\begin{tabular}[c]{@{}c@{}}95.86\\ ($\pm$ 1.80)\end{tabular}}} & \begin{tabular}[c]{@{}c@{}}95.69\\ ($\pm$ 1.99)\end{tabular} & \multicolumn{1}{c|}{\begin{tabular}[c]{@{}c@{}}88.24\\ ($\pm$ 3.74)\end{tabular}} & \textbf{\begin{tabular}[c]{@{}c@{}}89.84 \\ ($\pm$ 2.60)\end{tabular}} & \multicolumn{1}{c|}{\begin{tabular}[c]{@{}c@{}}92.64\\ ($\pm$ 3.37)\end{tabular}} & \textbf{\begin{tabular}[c]{@{}c@{}}93.51\\ ($\pm$ 2.52)\end{tabular}} \\ \hline
CLAM-MB & \multicolumn{1}{c|}{\textbf{\begin{tabular}[c]{@{}c@{}}96.53\\ ($\pm$ 1.73)\end{tabular}}} & \begin{tabular}[c]{@{}c@{}}96.27\\ ($\pm$ 1.51)\end{tabular} & \multicolumn{1}{c|}{\begin{tabular}[c]{@{}c@{}}88.74\\ ($\pm$ 2.64)\end{tabular}} & \textbf{\begin{tabular}[c]{@{}c@{}}89.72\\ ($\pm$ 4.03)\end{tabular}} & \multicolumn{1}{c|}{\begin{tabular}[c]{@{}c@{}}93.08\\ ($\pm$ 2.67)\end{tabular}} & \textbf{\begin{tabular}[c]{@{}c@{}}94.37\\ ($\pm$ 4.18)\end{tabular}} \\ \hline
TransMIL & \multicolumn{1}{c|}{\begin{tabular}[c]{@{}c@{}}95.52\\ ($\pm$ 2.76)\end{tabular}} & \textbf{\begin{tabular}[c]{@{}c@{}}96.85\\ ($\pm$ 1.20)\end{tabular}} & \multicolumn{1}{c|}{\textbf{\begin{tabular}[c]{@{}c@{}}90.58\\ ($\pm$ 1.93)\end{tabular}}} & \begin{tabular}[c]{@{}c@{}}89.36\\ ($\pm$ 2.30\end{tabular} & \multicolumn{1}{c|}{\begin{tabular}[c]{@{}c@{}}93.28\\ ($\pm$ 4.98)\end{tabular}} & \textbf{\begin{tabular}[c]{@{}c@{}}93.57\\ ($\pm$ 3.43)\end{tabular}} \\ \hline
\end{tabular}%
}
\end{table}


Table \ref{tab:structure} shows a comparison between ResNet50 and DenseNet121. We note that the structure of the generic pre-trained model also plays an important role in improving both predictive accuracy (by $1.1\%$ to $4.6\%$) and confidence (by $1.6\%$ to $9.6\%$). Since the DenseNet structure is utilized by KimiaNet, the gains made by using the DenseNet structure are incorporated in the results with KimiaNet. In combination, we believe the results provide evidence for the improved performance on model's confidence with domain-specific pre-training, irrespective of the CLAM or TransMIL models used for classification.

Classification of glioma types (i.e., astrocytoma, oligodendroglioma, and glioblastoma) is generally a difficult task; the current state-of-the-art performance is accuracy of 0.861 and AUC of 0.961 \cite{arpa2021}. Classification of glioma molecular subtypes is also challenging; for instance, the best binary classification performance of the IDH-gene molecular subtype is accuracy of 0.882, and AUC  of  0.931 \cite{scirep2020}. We test our models on the classification of both glioma types and molecular subtypes, which is a more challenging task, yet the models that use KimiaNet or DenseNet with TransMIL outperform this prior art.

\begin{table}[!htb]
\centering
\caption{Evaluation of the structure of feature extractor}
\label{tab:structure}
\resizebox{\columnwidth}{!}{%
\begin{tabular}{|c|cc|cc|cc|}
\hline
 & \multicolumn{2}{c|}{AUC} & \multicolumn{2}{c|}{Accuracy} & \multicolumn{2}{c|}{Confidence} \\ \hline
Models & \multicolumn{1}{c|}{ResNet} & DenseNet & \multicolumn{1}{c|}{ResNet} & DenseNet & \multicolumn{1}{c|}{ResNet} & DenseNet \\ \hline
CLAM-SB & \multicolumn{1}{c|}{\begin{tabular}[c]{@{}c@{}}95.43\\ ($\pm$ 1.35)\end{tabular}} & \textbf{\begin{tabular}[c]{@{}c@{}}95.86\\ ($\pm$ 1.80)\end{tabular}} & \multicolumn{1}{c|}{\begin{tabular}[c]{@{}c@{}}83.67 \\ ($\pm$ 4.23)\end{tabular}} & \textbf{\begin{tabular}[c]{@{}c@{}}88.24\\ ($\pm$ 3.74)\end{tabular}} & \multicolumn{1}{c|}{\begin{tabular}[c]{@{}c@{}}83.05\\ ($\pm$ 4.85)\end{tabular}} & \textbf{\begin{tabular}[c]{@{}c@{}}92.64\\ ($\pm$ 3.37)\end{tabular}} \\ \hline
CLAM-MB & \multicolumn{1}{c|}{\textbf{\begin{tabular}[c]{@{}c@{}}96.70\\ ($\pm$ 1.03)\end{tabular}}} & \begin{tabular}[c]{@{}c@{}}96.53\\ ($\pm$ 1.73)\end{tabular} & \multicolumn{1}{c|}{\begin{tabular}[c]{@{}c@{}}87.65\\ ($\pm$ 2.87)\end{tabular}} & \textbf{\begin{tabular}[c]{@{}c@{}}88.74\\ ($\pm$ 2.64)\end{tabular}} & \multicolumn{1}{c|}{\begin{tabular}[c]{@{}c@{}}89.07\\ ($\pm$ 4.18)\end{tabular}} & \textbf{\begin{tabular}[c]{@{}c@{}}93.08\\ ($\pm$ 2.67)\end{tabular}} \\ \hline
TransMIL & \multicolumn{1}{c|}{\begin{tabular}[c]{@{}c@{}}95.40\\ ($\pm$ 2.53)\end{tabular}} & \textbf{\begin{tabular}[c]{@{}c@{}}95.52\\ ($\pm$ 2.76)\end{tabular}} & \multicolumn{1}{c|}{\begin{tabular}[c]{@{}c@{}}88.18\\ ($\pm$ 3.53\end{tabular}} & \textbf{\begin{tabular}[c]{@{}c@{}}90.58\\ ($\pm$ 1.93)\end{tabular}} & \multicolumn{1}{c|}{\begin{tabular}[c]{@{}c@{}}91.69\\ ($\pm$ 3.92)\end{tabular}} & \textbf{\begin{tabular}[c]{@{}c@{}}93.28\\ ($\pm$ 4.98)\end{tabular}} \\ \hline
\end{tabular}%
}
\end{table}

\section{Conclusions}

This work is the first study that investigates the impact of domain-specific pre-training in WSI classification and uses confidence as a metric to evaluate the model's performance. We first consider the DenseNet model pre-trained on ImageNet and assess its accuracy, AUC and confidence when combined with CLAM and TransMIL to classify brain tumor WSIs. We then explore KimiaNet, a DenseNet121 model pre-trained on TCGA slides to tackle the same problem. We propose domain-specific pre-training of KimiaNet improves the classification accuracy, AUC and confidence of CLAM and TransMIL. The evaluation of our proposal on the DBTA dataset suggests that KimiaNet is a better feature extractor than generic DenseNet and ResNet, as it improves CLAM and TransMIL's confidence. Additionally, using TransMIL with KimiaNet achieves a new state-of-the-art performance of WSI-based glioma classification, which shows high clinical applicability of the model in assisting glioma diagnosis, particularly in predicting the molecular subtypes. This study also sheds light on the impact of the feature extractor's structure on the prediction performance. In our future work, we will investigate if advanced convolutional structures can further benefit multiple-instance learning models and improve their classification performance.

\bibliographystyle{IEEEtran}
\bibliography{references}


\end{document}